\pdfoutput=1

\documentclass[11pt]{article}
\makeatletter
\newif\ifacl@finalcopy
\acl@finalcopyfalse
\makeatother
\usepackage[review]{ACL2023}
\usepackage{CJKutf8}
\usepackage{graphicx}
\usepackage{times}
\usepackage{latexsym}
\usepackage{amsmath} 
\usepackage{tabularx} 
\usepackage{comment} 
\usepackage{fancyhdr}
\usepackage[T1]{fontenc}

\usepackage[utf8]{inputenc}

\usepackage{microtype}

\usepackage{inconsolata}
\usepackage{natbib}
\setcitestyle{numbers,square}
\title{AKEM: Aligning Knowledge Base to Queries with Ensemble Model for Entity Recognition and Linking}

 \author{Di Lu \and Zhongping Liang \and Caixia Yuan \and Xiaojie Wang \\
         Center for Intelligence Science and Technology \\ Beijing University of Posts and Telecommunications \\ Beijing, China\\
          \texttt{ludy.bupt@gmail.com }\\} 

\begin{document}
\maketitle

\begin{abstract}
This paper presents a novel approach to address the Entity Recognition and Linking Challenge at NLPCC 2015. The task involves extracting named entity mentions from short search queries and linking them to entities within a reference Chinese knowledge base. To tackle this problem, we first expand the existing knowledge base and utilize external knowledge to identify candidate entities, thereby improving the recall rate. Next, we extract features from the candidate entities and utilize Support Vector Regression and Multiple Additive Regression Tree as scoring functions to filter the results. Additionally, we apply rules to further refine the results and enhance precision. Our method is computationally efficient and achieves an F1 score of 0.535.
\end{abstract}

\section{Introduction}
\begin{CJK*}{UTF8}{gbsn}
The aim of Entity Recognition~\citep{cucchiarelli1998automatic,mikheev1999named} and Linking~\cite{artiles2007semeval,hachey2013evaluating} in Chinese Search Queries is to evaluate the current advancements of techniques in aligning named entities in short search queries to entities in a reference Chinese knowledge base~\cite{trochim2001research}. This task presents three main challenges. 

The first challenge involves the basic tasks of Chinese natural language processing, such as Chinese word segmentation~\cite{xue2003chinese}, POS tagging~\cite{voutilainen2003part}, and syntactic analysis~\cite{sportiche2013introduction}. 

The second challenge arises from the ambiguity and multiplicity of names: the same named entity string can occur in different contexts with different meanings (e.g., ``苹果" can refer to a type of fruit or Apple Inc.). Furthermore, the same named entity may be denoted using various strings, including abbreviations (e.g., ``中纪委") and full names (e.g.,  ``中共中央纪律检查委员会"). 

The final challenge is that the queried named entity may not exist in the knowledge base at all. This situation necessitates the ability to comprehend semantics and make inferences. This means the system needs to be capable of understanding the context and meaning of the query, and then use that understanding to infer possible matches or related entities, even if the exact entity queried is not present in the knowledge base.

The Entity Recognition and Linking Task can be divided into two sequential subtasks: 

1) Tagging mention, which is a typical word segmentation and named entity recognition (NER) task. 

2) Linking mentions to entities. Once mentions are recognized, the remaining difficulty lies in linking these mentions to entities and eliminating noisy candidate entities in cases where the mention can be linked with many candidate entities. 

In this paper, we propose AKEM(\textbf{A}ligning \textbf{K}nowledge Base to Queries with \textbf{E}nsemble \textbf{M}odel), which is designed to ensure high recall by initially recognizing all possible named entities at a relatively broad level. This is achieved by expanding the existing knowledge base and mining external knowledge to locate candidate entities. Subsequently, to ensure high precision, the method eliminates noisy entities through ensemble ranking and filtering rules.

\end{CJK*}
\section{Related Work}

The concept of Entity Linking was first proposed by Paul McNamee and Hoa Trang Dang~\cite{mcnamee2009overview}, marking a relatively new area of study within the field of natural language processing. The task primarily focuses on conducting entity extraction based on large-scale data and performing entity disambiguation.

Generally, a typical entity linking system comprises the following modules:
\begin{CJK*}{UTF8}{gbsn}
\begin{enumerate}
\item Query Extension Module: This leverages structured information from Wikipedia and reference background documents to extend the query and enrich its information content.
\item Candidate Entity Collection Generation Module: This identifies the set of candidate entities that may be linked to target entities.
\item Candidate Entity Collection Ranking Module: This employs an appropriate sorting algorithm to select the best matching candidate entity from the collection.
\item Non-KB-Entity Detection and Clustering Module: This detects entities that do not exist in the Knowledge Base (KB) and clusters their corresponding queries.
\end{enumerate}
\end{CJK*}

In the Entity Linking task, there isn't a significant difference in the query extension technology used across various systems. For entities that exist in the Knowledge Base (KB), the likelihood of a candidate entity matching the Gold Entity has reached a high level. Therefore, the critical step in Entity Linking is the candidate entity collection ranking module.

Common techniques employed in the candidate entity collection ranking module include:

\begin{enumerate}

\item  The Unsupervised Ranking Method: this method relies on the similarity between the background document of the query and the document of the candidate entity in the Knowledge Base as the basis for sorting. The target entity is more likely to have greater similarity. The advantage of this method is its simplicity and strong operability. However, this method is mostly based on an unannotated corpus, without utilizing the information in the training corpus. The lack of parameter learning and threshold adjustment in this method leads to inadequate information, making it less satisfactory.

\item The Supervised Ranking Method: this method treats the entity in the query and its corresponding entity to be linked in the KB as a pair for classification, and establishes a classification model. For instance, the Listwise Learning to Rank model employs the supervised ranking method~\cite{cao2007learning}. This method has the advantage of using the information in the training corpus and mining potential rules and information. However, it doesn't leverage some semantic information describing the target entity in the background document, thereby ignoring the role of the semantic information of target entities.

\item The Ranking Method Based on Graph Model: this method~\cite{radford2010document} a global optimization model using all entities in the document. This graph model method is adopted to improve the candidate collection sorting process. However, this method is rarely used in Entity Linking tasks and is left for further research.

\item The Ranking Method Based on Information Retrieval or Rules: Ne meskey et al.~\cite{nemeskey2010budapestacad} employ an information retrieval engine proposed by Daróczy~\cite{daroczy2010jumas} to sort the candidate entity set. This method retrieves the most relevant entity document of KB in Wikipedia. However, as there are many other entities unrelated to the querying entity, using the whole document as a retrieval condition adds a lot of noisy information to the retrieval. Therefore, this method has significant limitations and does not perform well. Gao et al.~\cite{gao2010pris} sort the candidate entity set using a rule-based method. In their system, they design specific rules for three different types of entities: Pre, Org, and Gpe. However, due to the variability of test corpora, these established rules may not necessarily apply universally. It is unpredictable whether these established rules will enhance performance when processing different test corpora. Therefore, this method has its limitations.

\end{enumerate}

In this paper, we introduce a novel candidate entity collection ranking module that integrates multiple methods. Our module primarily employs a supervised learning method based on Support Vector Regression and Multiple Additive Regression Trees. Concurrently, it also utilizes the ranking method based on information retrieval and rules. Our method is computationally efficient and achieves an F1 score of 0.535.

\section{Framework}
\begin{figure*}[t]
  \centering
  \includegraphics[width=0.99\textwidth]{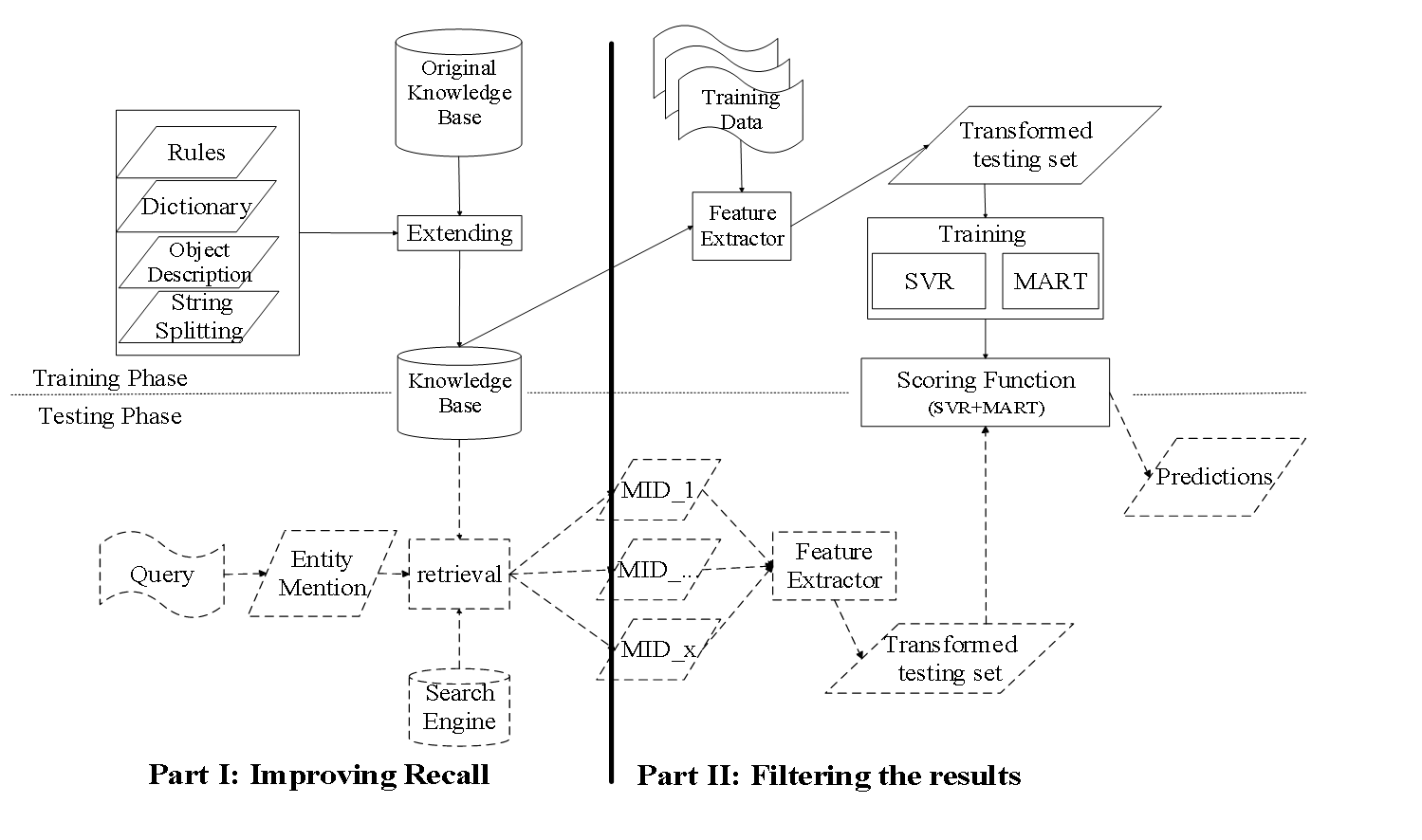}
  \caption{Flowchart of AKEM. Initially, in Part I, we expand the Knowledge Base and conduct a search for a broader set of candidate entities. Following this, in Part II, we refine these results using a Scoring Function.}
  \label{fig:model}
\end{figure*}
The overall processing flow of our system is shown in Figure~\ref{fig:model}. As shown in Figure~\ref{fig:model}, AKEM has two phases. The first phase is to improve the recall rate by extending the existing Chinese knowledge base and using search engine. The second phase is to remove noisy candidate entities by using SVR-MART joint scoring functions and rule-based filter.

Details of the three parts are given one by one in followings.
\subsection{Improving Recall}

In order to improve recall rate, our model need to recognize as many candidate entities as possible.
\subsubsection{Extending Knowledge Base}
NLPCC 2015~\cite{feng2015overview} provides us with a reference Chinese Knowledge Base (KB). To fully leverage this KB, we need to expand it using the following methods:
\begin{CJK*}{UTF8}{gbsn}
\begin{enumerate}

\item Processing English Entities: The English entities in the Knowledge Base are in a unique format. For example, ``Microsoft\_Word'' or ``As\_Long\_As\_You\_Love\_Me''. Names in the Knowledge Base start with capital letters and are connected by underscores, complicating direct matching. Therefore, we need to remove underscores and convert the names to lowercase.
\item Removing Brackets: To identify more named entities in the KB, we remove brackets in entity names and split English names in the KB. For example, new entity names like ``李安'' and ``李安工程师'' are mapped to the entity ``李安\_(工程师)'' in the KB by removing brackets. ``奥巴马'' is mapped to the entity ``贝拉克·奥巴马'' by splitting the English name.
\item Establishing a Place Directory by Heuristic Inference: All entity names of places in the KB are full names, like ``北京市'' and ``河北省''. We can infer whether a place is a city, county, province, village, or a county based on the last word of the entity name.
\item Entity Name Extension by Heuristic Method: Many entity names are not common. For instance, ``白马银花'' is a nickname for ``香港杜鹃''. Some entities in the KB have descriptions that lead us to find nicknames of entities. Therefore, regular expressions can be used to extract these nicknames. For example, the regular expression ``别称(.?)'' or ``别名 (.?)'' can extract a nickname of an entity from the object description of an entity in the KB. 
\end{enumerate}
\end{CJK*}

\subsubsection{Searching For Candidate Entities}
An entity ID may correspond to multiple mentions. When a query is input, our method manages Chinese word segmentation. We search for each mention resulting from the word segmentation using the Baidu search engine, selecting the top 10 search results. We then match each of these search results with the entity names in the extended Knowledge Base. Each successful match and its corresponding entity ID are saved. This approach allows us to recognize a greater number of candidate entities, thereby improving the recall rate.

\subsection{Filtering the Results}

After identifying as many candidate entities as possible to enhance the recall rate, it becomes necessary to eliminate noisy entities to improve the precision rate. Initially, we extract features using a template method and our defined formulas. Subsequently, we employ scoring functions based on Support Vector Regression and Multiple Additive Regression Trees to filter out noisy entities. Finally, we apply specific rules to select the appropriate entities. In this way, we first utilize a statistical method followed by a rule-based method to effectively filter out noisy entities.

\subsubsection{Feature Extraction}
We extract features for each mention-candidate entity pair, employing the following selection process for these features.

\begin{enumerate}

\item The similarity between the query and the object description of the candidate entity is considered. We define this similarity, denoted as $Similarity_1(query, object)$, as follows:
\begin{equation}
\label{eq:Similarity1}
Similarity_1(query, object) = \frac{2 \cdot c_1}{l_1 + l_2}
\end{equation}\\
In the aforementioned function, $c_1$ represents the count of identical characters between the query and the object description of the candidate entity. $l_1$ denotes the character count in the query, while $l_2$ signifies the character count in the object description.
\item Whether the mention of candidate entity exclusively contains numerical characters(0-9). 
\item Whether the mention of candidate entity contains numerical characters (0-9).
\item Whether the mention of candidate entity exclusively contains letters, either lower case (a-z) or upper case (A-Z).
\item Whether the mention of entity candidate contains letters, either lower case (a-z) or upper case (A-Z).
\item Whether the candidate entity name is a substring of the query.
\item Whether the query is a substring of the candidate entity name.
\item The maximum similarity between each word in the object description of the current mention and other mentions in a query is calculated. Word embeddings are used to represent a mention or a word. The Word2Vec model is employed to train the word embedding using the existing Knowledge Base provided by NLPCC 2015 as training data. We use cosine similarity to calculate the similarity between a word and a mention.
\begin{eqnarray}
\begin{gathered}
\forall k \neq i, \forall j,  \\
f_8(\mathbf{m}_{i}) = \max(Similarity_2(\mathbf{w}_{ij} \cdot \mathbf{m}_{k})),  \\
Similarity_2(\mathbf{w}_{ij}, \mathbf{m}_{k}) = \frac{\mathbf{w}_{ij} \cdot \mathbf{m}_{k}}{||\mathbf{w}_{ij}|| + ||\mathbf{m}_{k}||}.
\end{gathered}
\end{eqnarray}

\begin{table*}[htb]
\centering
\begin{tabular}{p{2cm} p{2cm} p{2cm} p{2cm} p{2cm}}
\hline
\textbf{Team} & \textbf{Precision} & \textbf{Recall} & \textbf{Link-F1} & \textbf{Average-F1} \\
\hline
Our team & 0.480 & \textbf{0.656} & 0.555 & \textbf{0.535} \\
Average & 0.450 & 0.497 & 0.458 & 0.423 \\
\hline
\end{tabular}
\caption{\label{full-result}
Evaluation results in the NLPCC 2015 NEL task.
}
\end{table*}

In the function above, $f_8(\mathbf{m}_{i})$ represents the feature value of this dimension for the i\textit{th} mention in a query. ${\mathbf{w}_{ij}}$ refers to the word embedding of the j\textit{th} word in the object description of the i\textit{th} mention. $\mathbf{m}_{k}$ is the word embedding of the k\textit{th} mention in the query. If we are unable to obtain the word embedding of a word, the similarity value between this word and any other word is set to a default value of -1.

\item Whether there exists other mention in the query which is a substring of object description of the candidate entity. 
\end{enumerate}
\subsubsection{Statistical-Based Filtering}
After obtaining the candidate entities of a query and the feature vector for each candidate entity, our method utilizes statistical methods to construct a scoring function that filters out noisy entities. We carry out these processes sequentially:

\noindent\textbf{Training Set.} The training set is derived from the 159 queries and their corresponding entities provided by NLPCC 2015. We employ our method to identify candidate entities for each query and compute the feature vector for each candidate entity. If a candidate entity $e_i$ matches the standard entity list of the query, we tag $e_i$ with 1. Otherwise, we tag $e_i$ with 0.\\
\begin{equation*}
y = \begin{cases}
1, & \text{if candidate entity $e_i$ in $s$} \\
0, & \text{otherwise}
\end{cases}\\
\end{equation*}
In this manner, we can construct the training set:\\ ${\{(x_1,y_1), \ldots, (x_n, y_n)\} \subseteq X \times [0.0, 1.0]}$. Here, ${x_i}$ represents the feature vector of ${e_i}$, and ${y_i}$ represents the tag of ${e_i}$. 

\noindent\textbf{Support Vector Regression(SVR).} Drawing on the principles of Support Vector Machines~\cite{hearst1998support, noble2006support}, Support Vector Regression~\cite{drucker1996support, smola2004tutorial,awad2015support} is a well-established method for identification, estimation, and prediction. We utilize the tagged training set to train a SVR model. Subsequently, this model is applied to score each candidate entity, facilitating an effective filtering process. Let $f_{SVR}(\mathbf{x_e})$ represent the score assigned by the SVR model to a candidate entity $e$, and $\mathbf{x_e}$ be the feature vector of the candidate entity $e$. Then, $f_{\text{SVR}}(\mathbf{x_e}) \in [0.0, 1.0]$. 

\noindent\textbf{Multiple Additive Regression Tree(MART).}   \textit{Multiple Additive Regression Trees}~\citep{friedman2003multiple, vinayak2015dart} is an ensemble model of boosted regression trees, known for delivering high prediction accuracy across diverse tasks and widely used in practice. We train a MART model using the tagged training set. This model is then employed to score each candidate entity that is to be filtered. Let $f_{MART}(\mathbf{x_e})$ represent the score assigned by the MART model to a candidate entity $e$. Then, $f_{\text{MART}}(\mathbf{x_e}) \in [0.0, 1.0]$. 

\noindent\textbf{Scoring function.} We define the scoring function ${Score}(\mathbf{x_e})$ as follow: \\
\begin{equation}
\label{eq:scoring}
Score(\mathbf{x_e}) = \frac{f_{SVR}(\mathbf{x_e}) + f_{MART}(\mathbf{x_e})}{2} .
\end{equation}\\
In the scoring function mentioned above, $\mathbf{x_e}$ represents the feature vector of the candidate entity $e$, and ${Score}(\mathbf{x_e}) \in [0.0, 1.0]$. 

\noindent\textbf{Filtering.} Each mention identified in a query may correspond to one or more candidate entities. To eliminate noisy entities, we use a threshold $\alpha$ to filter the candidate entities for each mention. Let $L$ represent the list of candidate entities for a mention $m$. We sort the candidate entities in $L$ in descending order based on their scores. For each mention in a query, we select the top $k$ candidate entities (if the number of candidate entities is less than $k$, we select all candidate entities). This results in a candidate list $L_k$ for further filtering.
Next, we select the top $l$ candidates in $L_k$ for each mention. Here, $l \leq k$. We determine $l$ using the following equation:
\begin{equation}
\label{eq:Filtering}
l = \arg\min_{i \leq k} \sum_{i=1}^{k} Score(e_i) \geq \alpha
\end{equation}

In the equation above, we begin by adding up the scores of candidate entities in $L_k$, starting with the highest score. Once the sum reaches the threshold $\alpha$, we retain the candidate entities that have been included in the calculation and remove the others. If  $\sum_{i=1}^{k} Score(e_i) < \alpha$, it indicates that all candidate entities in $L_k$ are unqualified, and we therefore remove all candidate entities in $L_k$.

\subsubsection{Rule-Based Filtering}
The final stage of our method relies on the results filtered by the scoring function. We implement the following rules to eliminate noisy entities:
\begin {enumerate}
\item If an English string is divided into several candidate entities, these candidates are removed.

\item Single Chinese character candidates are filtered out.

\item If multiple mentions in a query link to a single candidate entity, we use formula (\ref{eq:Similarity1}) to select the mention most similar to the candidate entity as the entity mention.
\item If a mention in a query can link to more than one candidate entity, and one of the candidates is identical to the mention, we select that candidate entity and remove the others.
\end{enumerate}
\section{Experiments}

We utilize 159 labeled queries, provided by NLPCC 2015, as training data to train the SVR and MART models, which we use as scoring functions. We then apply our model to a set of 3849 queries in order to evaluate our method. We set threshold $\alpha=0.3$ and  $k=3$. The performance of our method on the NLPCC 2015 evaluation dataset is presented in Table~\ref{full-result}. 

In the evaluation results, AKEM significantly outperforms the average scores in terms of Precision, Recall, Link-F1, and Average-F1. Specifically, our method exceeds the average Recall score by 15.9\% and surpasses the average Average-F1 score by 11.2\%. In terms of final rankings, our method secures the fourth position among all participating teams in both Recall rate and Link-F1 score, as well as Average-F1 score. This demonstrates the competitive performance of our approach in the context of the other teams.

\section{Conclusion}
We aim to construct an experimental framework, AKEM, using a reference Chinese knowledge base. This framework is developed and applied to the Entity Recognition and Linking task of NLPCC 2015. Initially, the framework recognizes as many candidate entities as possible by extending the knowledge base and utilizing a search engine, thereby increasing the recall rate. Subsequently, noisy candidate entities are eliminated through feature extraction, SVR-MART filtering, and rule-based filtering.

Our AKEM method exhibits a relatively strong performance in the Entity Recognition and Linking task during NLPCC 2015. It ranks fourth among all teams in terms of recall rate, Link-F1 score, and Average-F1 score.

\section*{Acknowledgements}
This paper was completed during the NLPCC 2015 Shared Task: Entity Recognition and Linking in Search Queries. As a participating team, we competed in the Entity Linking task of NLPCC 2015. The evaluation results were officially assessed and announced by the NLPCC 2015 committee, and the task data copyright belongs to NLPCC 2015.

\bibliography{custum}

\end{document}